\documentclass[acmsmall,nonacm]{acmart}

\usepackage{multirow}
\usepackage{subfigure}

\AtBeginDocument{%
  }

\renewcommand\footnotetextcopyrightpermission[1]{} 
\makeatletter
\let\@authorsaddresses\@empty
\makeatother

\begin{document}

\title{Collaborative Learning of On-Device Small Model and Cloud-Based Large Model: Advances and Future Directions}

\author{Chaoyue Niu}
\authornote{Chaoyue Niu is the corresponding author (rvince@sjtu.edu.cn).}
\email{rvince@sjtu.edu.cn}
\orcid{0000-0002-1650-4233}
\affiliation{%
  \institution{Shanghai Jiao Tong University}
  \city{Shanghai}
  \country{China}
}

\author{Yucheng Ding}
\affiliation{%
  \institution{Shanghai Jiao Tong University}
  \city{Shanghai}
  \country{China}
}
\email{yc.ding@sjtu.edu.cn}

\author{Junhui Lu}
\affiliation{%
  \institution{Shanghai Jiao Tong University}
  \city{Shanghai}
  \country{China}
}
\email{ljh123@sjtu.edu.cn}

\author{Zhengxiang Huang}
\affiliation{%
  \institution{Shanghai Jiao Tong University}
  \city{Shanghai}
  \country{China}
}
\email{huangzhengxiang@sjtu.edu.cn}

\author{Hang Zeng}
\affiliation{%
  \institution{Shanghai Jiao Tong University}
  \city{Shanghai}
  \country{China}
}
\email{nidhogg@sjtu.edu.cn}

\author{Yutong Dai}
\affiliation{%
  \institution{Shanghai Jiao Tong University}
  \city{Shanghai}
  \country{China}
}
\email{daiyutong@sjtu.edu.cn}

\author{Xuezhen Tu}
\affiliation{%
  \institution{Shanghai Jiao Tong University}
  \city{Shanghai}
  \country{China}
}
\email{xuezhentu@sjtu.edu.cn}

\author{Chengfei Lv}
\affiliation{%
  \institution{Alibaba Group}
  \city{Hangzhou}
  \state{Zhejiang}
  \country{China}
}
\email{chengfei.lcf@alibaba-inc.com}

\author{Fan Wu}
\affiliation{%
  \institution{Shanghai Jiao Tong University}
  \city{Shanghai}
  \country{China}
}
\email{wu-fan@sjtu.edu.cn}

\author{Guihai Chen}
\affiliation{%
  \institution{Shanghai Jiao Tong University}
  \city{Shanghai}
  \country{China}
}
\email{gchen@cs.sjtu.edu.cn}

\renewcommand{\shortauthors}{Niu et al.}

\begin{abstract}
The conventional cloud-based large model learning framework is increasingly constrained by latency, cost, personalization, and privacy concerns. In this survey, we explore an emerging paradigm: collaborative learning between on-device small model and cloud-based large model, which promises low-latency, cost-efficient, and personalized intelligent services while preserving user privacy. We provide a comprehensive review across hardware, system, algorithm, and application layers. At each layer, we summarize key problems and recent advances from both academia and industry. In particular, we categorize collaboration algorithms into data-based, feature-based, and parameter-based frameworks. We also review publicly available datasets and evaluation metrics with user-level or device-level consideration tailored to collaborative learning settings. We further highlight real-world deployments, ranging from recommender systems and mobile livestreaming to personal intelligent assistants. We finally point out open research directions to guide future development in this rapidly evolving field. 
\end{abstract}



\maketitle

\section{Introduction}

\begin{figure}[!t]
    \centering
    \subfigure[Cloud-Based Large Model Learning Paradigm]{\label{fig:cloud:large}\includegraphics[width=\columnwidth]{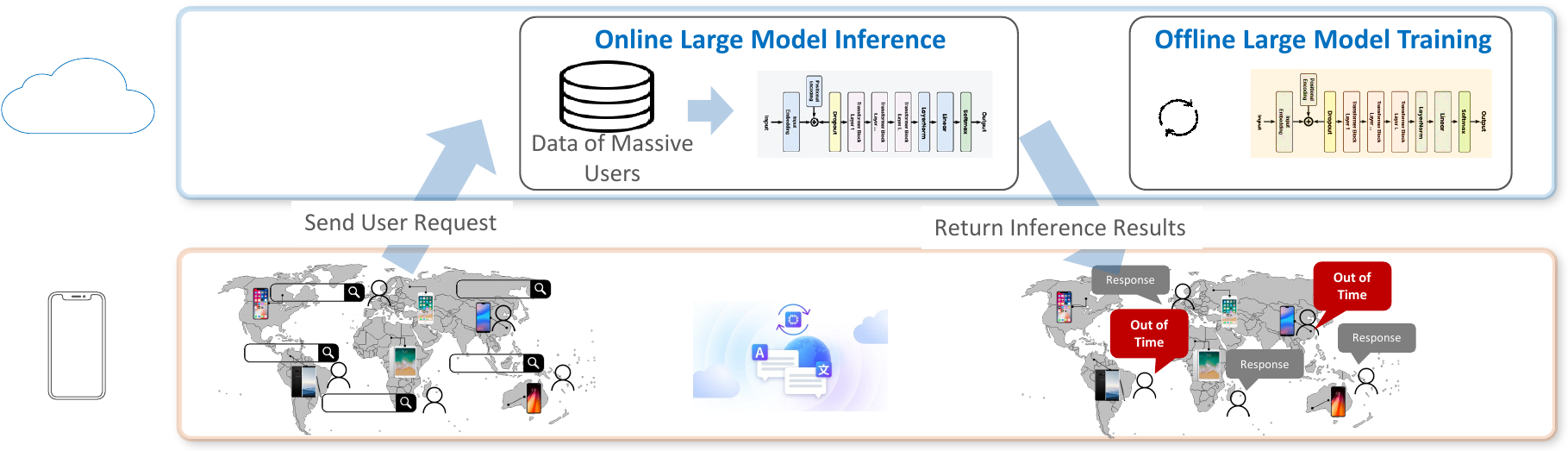}}
    \hfill
    \subfigure[On-Device Small Model and Cloud-Based Large Model Collaborative Learning Paradigm]{\label{fig:device:cloud}\includegraphics[width=\columnwidth]{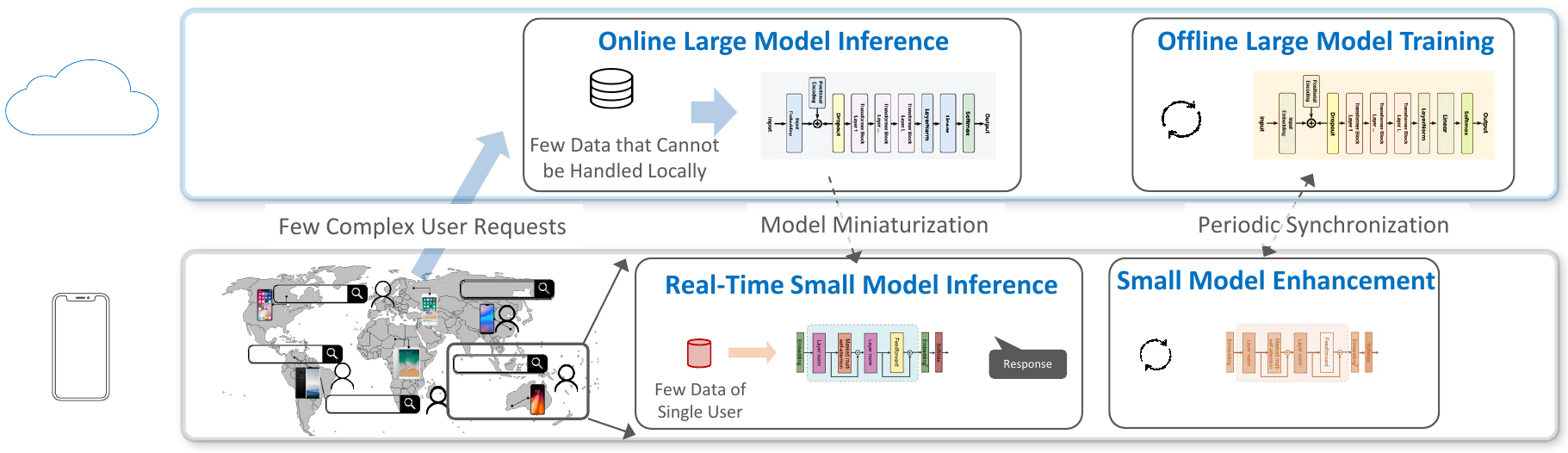}}
    \caption{Learning paradigms comparison.}
    \Description{Please see caption.}
\end{figure}

To deliver intelligent services to millions or even billions of mobile device users, as shown in Fig. \ref{fig:cloud:large}, the mainstream paradigm relies on sending raw data from mobile devices to the cloud, where the large model processes the data and return results. In addition to traditional discriminative models, the past two years have seen remarkable advances in large generative models, such as GPT, Qwen, LLaMA, DeepSeek, and Stable Diffusion, driven by increasing volumes of training data, more powerful computational infrastructure, and continual improvements in learning algorithms and model architectures. These large models are capable of generating high-quality and highly creative contents and have also demonstrated strong performance on complex reasoning tasks, addressing the long-standing fragmentation of traditional task-specific models and marking a significant step toward artificial general intelligence.

However, under the cloud-based large model learning framework, the prevalent reliance on cloud servers for training and inference by centralizing both data and computation introduces several critical bottlenecks when serving massive mobile devices across various vertical applications. The first bottleneck is {\em high response latency}. The network latency between each mobile device and the cloud, combined with the request processing latency on the cloud with the large model, typically amounts to several seconds. This level of delay is unacceptable for many real-time interactive applications, often on the order of hundreds or even tens of milliseconds. Additionally, during peak concurrent user requests, response latency increases significantly. In scenarios with poor mobile network conditions, cloud interface access limits, or cloud service failures, the cloud-based large model service becomes unavailable. The second bottleneck is {\em high cost and heavy load.} On the mobile device side, uploading raw data can result in significant cellular data usage, particularly when Wi-Fi is unavailable. Some large generative models are often offered as paid services. On the cloud side, receiving and storing vast amounts of raw data from a large number of mobile devices, processing data with large models with billions or even trillions of parameters, and returning results in time, inevitably cause high overhead. The overall cost increases with longer service durations and a growing user base. For example, Vivo estimates that for 300 million mobile device users making 10 cloud-based generative model requests daily, the cost could reach approximately 30 million yuan per day, totaling over 9 billion yuan annually \cite{link:device:llm:tencent}. The third bottleneck is {\em insufficient personalization}. A single large model deployed in the cloud, trained on global data rather than individual user data, struggles to deliver services tailored to each user's unique characteristics and needs. The further development of large generative models also faces the challenge of a dwindling supply of high-quality public data. To enable better personalization, it becomes increasingly important for the cloud-based large model to perceive and leverage private user data residing on mobile devices. The fourth bottleneck is {\em high data security and privacy risk}. Uploading the raw data with sensitive user information raises serious security and privacy concerns of users. Storing and processing raw data on the cloud may suffer from the risk of data breach.

To break the bottlenecks of the cloud-based large model learning framework, academia and industry explore deploying small model on mobile devices and letting the cloud-based large model and the on-device small model collaboratively execute learning tasks, as depicted in Fig. \ref{fig:device:cloud}. The cloud-based large model can transfer essential domain-specific knowledge to the on-device small model, enabling close model collaboration for real-time inference tailored to each mobile device user. Such a new distributed, self-service paradigm leverages the natural advantages of mobile devices being at application scenarios, users, and data sources, thereby reducing latency and communication cost, alleviating the cloud-side load by exploiting mobile device resources, and keeping private data on local devices to align with emerging data protection and regulatory requirements. Moreover, the on-device small model can continuously adapt to the user's context by leveraging the stream of user-specific data, improving service personalization and accuracy to meet diverse and evolving needs. The on-device small model can also feed back execution results, fresh insensitive samples, personalized features, and updated parameters to the cloud-based large model, supporting continual improvement. This forms a virtuous cycle of co-evolution, laying the foundation for a long-term vision of ubiquitous, collaborative intelligence. At the end of 2021, Alibaba's DAMO Academy identified the co-evolution of cloud-based large model and on-device small model as one of the top ten technology trends for 2022 \cite{link:damo:tech:trend}. In May 2023, Qualcomm released a white paper titled ``The Future of AI is Hybrid'' \cite{link:Qualcomm:white:paper:part1, link:Qualcomm:white:paper:part2}. Qualcomm envisioned that generative artificial intelligence (AI) workloads must be distributed and coordinated between the cloud and mobile devices, rather than processing on the cloud alone, to offer benefits with regards to cost, energy, performance, privacy, security, and personalization at a global scale, just like traditional computing evolved from mainframes and thin clients to today’s mix of cloud and mobile devices.

In this survey, we first overview the framework from different layers in Section \ref{sec:framework}; then outline key problems in Section \ref{sec:key:problem}; further introduce representative advances from both academia and industry in Section \ref{sec:advance}; and finally point out future directions in Section \ref{sec:future:dir}.

\section{Framework Overview}\label{sec:framework}

\begin{figure}[!t]
  \centering
  \includegraphics[width=\columnwidth]{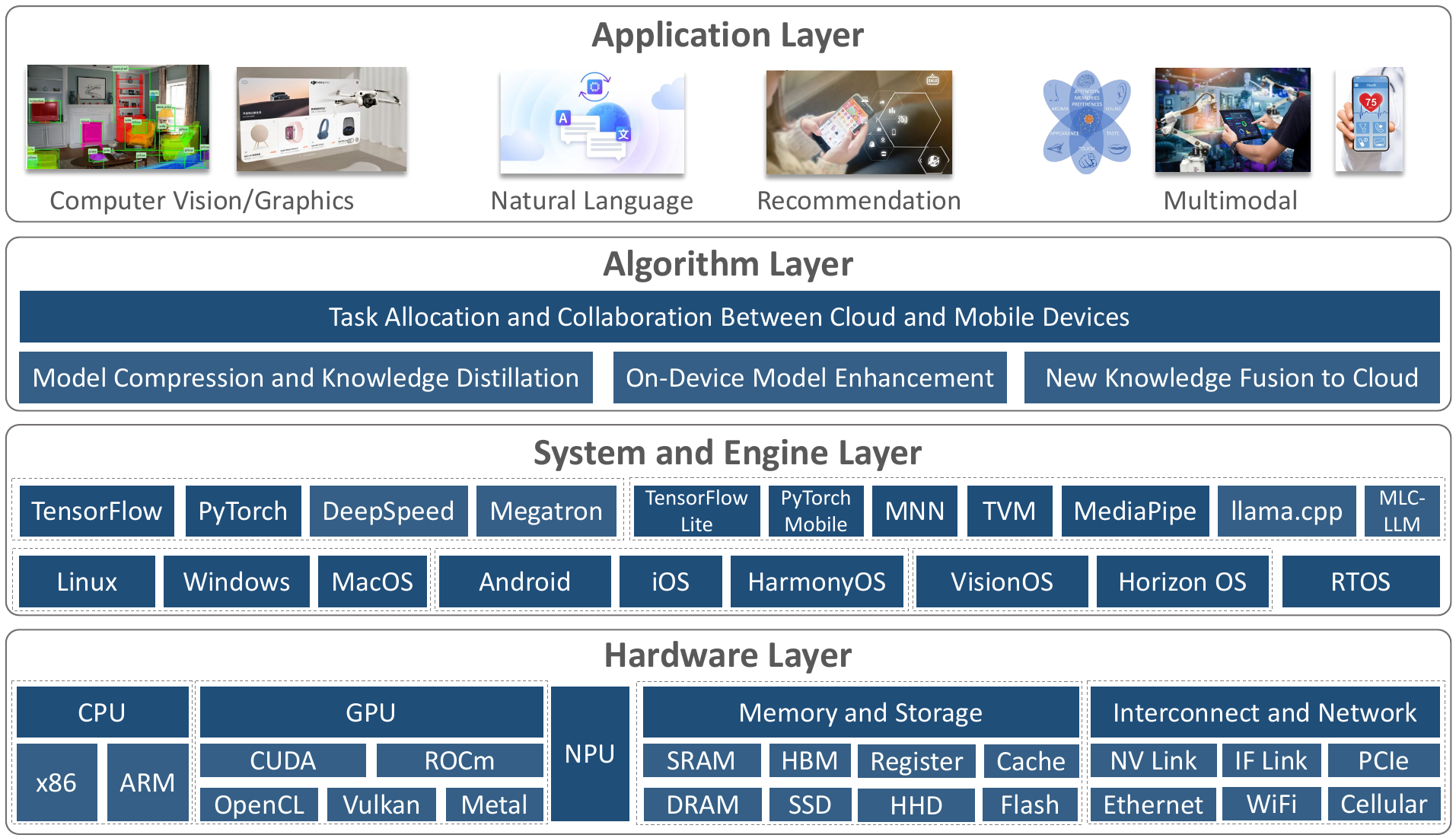}
  \caption{Overall framework.}\label{fig:framework}
  \Description{See the caption.}
\end{figure}

We introduce the collaborative learning framework of on-device small model and cloud-based large model in a bottom-up way. The framework encompasses the hardware layer, the system and deep learning engine layer, the algorithm layer, and the application layer. An overview of the framework is illustrated in Fig. \ref{fig:framework}.

\subsection{Hardware Layer}\label{sec:framework:hardware} 

\begin{table}[!t]
\caption{Typical hardware specifications for cloud servers and mobile devices.}\label{tab:hardware}
\centering
\resizebox{\columnwidth}{!}{
\begin{tabular}{l|l|l|l|l|l}
\toprule
Platform & Device   & Computation & Memory & Interconnect/Network & Power \\\midrule
& \multirow{2}{*}{NVIDIA H100 SXM} &  FP32: 67 TFLOPS & \multirow{2}{*}{80GB (3.35TB/s)} & NVLink: 900GB/s;  & \multirow{2}{*}{700W}\\
        & & INT8: 3958 TOPS & & PCIe 5.0 $\times$ 16: 128GB/s & \\\cline{2-6}
         & \multirow{2}{*}{NVIDIA RTX 4090} & FP32: 82.6 TFLOPS & \multirow{2}{*}{24GB (1008GB/s)} & \multirow{2}{*}{PCIe 4.0 $\times$ 16: 64GB/s} & \multirow{2}{*}{450W}\\
         & & INT8: 660.6 TOPS & & &\\\cline{2-6}
         & \multirow{2}{*}{AMD Instinct MI250X} & FP32: 47.9 TFLOPS & \multirow{2}{*}{128GB (3.2TB/s)} & Infinity Fabric Link: 100GB/s &  \multirow{2}{*}{500W}\\
Cloud    &                     & INT8: 383 TOPS & & PCIe 4.0 $\times$ 16: 64GB/s & \\\cline{2-6}
Server  & \multirow{2}{*}{Intel Xeon Platinum 8480C} & 56 Cores; & \multirow{2}{*}{4TB (307.2 GB/s)} & Intel UPI $\times$ 4: 128GB/s & \multirow{2}{*}{350W} \\
         &  & 2 GHz (Base); 3.9GHz (Turbo) & & PCIe 5.0 $\times$ 80: 640GB/s & \\\cline{2-6}
         & \multirow{2}{*}{AMD EPYC 7713P} & 64 Cores & \multirow{2}{*}{4TB (204.795GB/s)} & \multirow{2}{*}{PCIe 4.0 $\times$ 128: 512GB/s} & \multirow{2}{*}{225W} \\
         &                & 2 GHz (Base); 3.67 GHz (Boost) & & &\\\cline{2-6}
         & \multirow{2}{*}{Google TPU v5p} & 2 Cores Per Chip & \multirow{2}{*}{95GB (2765GB/s)} & \multirow{2}{*}{600GB/s} & \multirow{2}{*}{400W}\\
         &                & INT8: 918 TOPS Per Chip & & \\\midrule
         & Apple iPhone 16 Pro & 6-Core CPU (2.42GHz--4.05GHz); & \multirow{2}{*}{8GB (60GB/s)} & \multirow{2}{*}{5G (Snapdragon X75)} & \multirow{2}{*}{13.942Wh}\\
         & (A18 Pro) &   6-Core GPU; 16-Core NPU (35 TOPS) & & \multirow{2}{*}{Up: 3.5Gbps, Down: 10Gbps} & \\\cline{2-4}\cline{6-6}
    Smart & Samsung Galaxy S24 Ultra & 8-Core CPU (2.2GHz--3.39GHz) & \multirow{2}{*}{12GB (77GB/s)} & \multirow{2}{*}{/ WiFi 7 (40Gbps)} & \multirow{2}{*}{18.5Wh} \\
    Phone & (Snapdragon 8 Gen 3) & Adreno 750 GPU; Hexagon NPU &  & &\\\cline{2-6}
         & Huawei Mate 60 Pro & 8-Core CPU (1.53GHz--2.62GHz) & \multirow{2}{*}{12GB (44GB/s)} & 5G (Balong 5000) Up: 2.5Gps, & \multirow{2}{*}{18.5Wh} \\
         & (Kirin 9000S) & Maleoon 910 GPU; Da Vinci NPU & &  Down: 4.6Gps / WiFi 6 (2.5Gbps) & \\\midrule
      & Apple Vision Pro & 8-Core CPU (2.42GHz--3.49GHz); & \multirow{2}{*}{16GB (256GB/s)} & \multirow{2}{*}{WiFi 6 (2.5Gbps)} & \multirow{2}{*}{35.9Wh}\\
    XR  & (M2 + R2) & 10-Core GPU; 16-Core NPU & & &\\\cline{2-6}
    Device  & Meta Quest 3 & 8-Core CPU (2.0GHz--3.19GHz) & \multirow{2}{*}{8GB (25.6 GB/s)} & \multirow{2}{*}{WiFi 6E (3.6Gbps)} & \multirow{2}{*}{18.88Wh} \\
      & (Snapdragon XR2 Gen 2) & Adreno 740 GPU & & & \\\midrule
    & \multirow{2}{*}{NVIDIA Jetson AGX Orin} & INT8: 275 TOPS; 2048-Core GPU; & \multirow{2}{*}{64GB (204.8GB/s)} & \multirow{2}{*}{10GbE / GbE} & 15W  --  \\
    &                        & 12-Core CPU (2.2GHz); 2 NVDLA & & & 60W\\\cline{2-6}
  Embedded  & \multirow{2}{*}{NVIDIA Jetson TX2} & INT8: 1.33 TFLOPS; 256-Core GPU; & \multirow{2}{*}{8GB (59.7 GB/s)} & \multirow{2}{*}{GbE / WLAN} & 7.5W --\\
  Device & & 2-Core CPU (2.2GHz + 2GHz) &&& 15W\\\cline{2-6}
    & \multirow{2}{*}{NVIDIA Jetson Nano} & INT8: 472 GFLOPS; 128-Core GPU; & \multirow{2}{*}{4GB (25.6 GB/s)} & \multirow{2}{*}{GbE} & 5W --\\
    & & 2-Core CPU (1.43GHz) &&& 10W\\\midrule
 Micro-  & STM32 F746NG & 462 DMIPS & 320KB & 10/100 Mbps Ethernet MAC & 360mW\\\cline{2-6}
 Controller   & TI AM2434 & 2500 DMIPS & 2MB & 10/100/1000 Mbps RGMII & 2W \\
\bottomrule
\end{tabular}
}
\end{table}

At the bottom hardware layer, the resources of computation, memory, storage, and communication are provided to support model execution. The cloud server is resource-rich, whereas mobile device is resource-constrained. Typical hardware specifications for cloud server, smartphones, extended reality (XR) devices, embedded devices, and microcontrollers are summarized in Table \ref{tab:hardware}. 

The cloud infrastructure is typically provisioned with a cluster of high-performance CPUs, GPUs, or TPUs, complemented by high-bandwidth memory and interconnects, to facilitate large model training and inference. For example, Llama 3 405B is trained on up to 16K NVIDIA H100 GPUs using Meta’s Grand Teton AI server platform \cite{arxiv:llama3}, where each server is equipped with 8 GPUs and 2 CPUs, and the INT8 performance of one NVIDIA H100 reaches 3958 TOPS. The memory hierarchy for GPU includes static random-access memory (SRAM) and high-bandwidth memory (HBM). The memory and storage hierarchy for CPU comprises multiple levels, including registers, L1/L2/L3 caches, main memory utilizing dynamic random-access memory (DRAM), solid-state drives (SSD), and hard disk drives (HDD).  As moving down the hierarchy, storage capacity increases, whereas speed and bandwidth decrease. For example, NVIDIA A100 features 20MB of SRAM with speeds up to 19 TB/s and 40GB of HBM with speeds up to 1.5 TB/s \cite{proc:nips22:flashattention}. Interconnects for GPUs and CPUs on a server utilize point-to-point links to achieve high bandwidth, such as NVLink 4.0 of 900 GB/s per link, Intel Ultra Path Interconnect (UPI) 4.0 of 32 GB/s per link, and PCIe 5.0 of 8 GB/s per lane. To connect multiple servers in a data center environment, the principal networking technologies include RDMA over Converged Ethernet (RoCE) and InfiniBand.

Mobile devices are now equipped with advanced low-power CPUs, GPUs, and NPUs, which, due to significant improvements over the past decade, can handle efficient model inference and training workloads. Mobile devices can also connect to the cloud through wireless networks, such as WiFi and cellular, to collaborate on complex learning tasks. (1) From {\em computation resources}, modern smartphones and XR devices are generally equipped with CPUs featuring 6 to 8 cores, operating at frequencies between 2 GHz and 3 GHz. Embedded devices tend to excel in GPU computing power. For example, NVIDIA Jetson AGX Orin boasts a 2048-core GPU, achieving an INT8 performance of up to 275 TOPS. Microcontrollers, in contrast, are computationally constrained, with their Dhrystone million instructions per second (DMIPS) number typically ranging from a few hundred to several thousand; (2) from {\em running memory}, smartphones typically come equipped with 6GB to 12GB, while XR devices offer a bit more, ranging from 8GB to 16GB. Embedded devices have a broader spectrum of memory capacities, from 4GB on NVIDIA Jetson Nano to 64GB on NVIDIA Jetson AGX Orin. Microcontrollers are, instead, quite memory-limited, with only hundreds of KB to a few MB; (3) from {\em power}, the battery capacity of smartphones and XR devices is typically measured in tens of watt-hours (Wh) \footnote{Battery capacity, rather than instantaneous power consumption, is the more relevant metric for user-end devices like smartphones and XR devices, as they operate unplugged for most of their use. Consequently, these devices are typically specified in Wh, which reflect how long they can run on a single charge. In contrast, devices that stay plugged in are more commonly characterized by their power draw in W, since usage time is less of a concern.}, which allows for several hours of usage. This is because their power consumption usually ranges from 1.5 to 10 watts (W), depending on factors such as workload intensity, frame rate, and power-saving settings. The power consumption of embedded devices varies from a few W to several tens of W. In contrast, microcontrollers have a power consumption ranging from hundreds of mW to a few W; and (4) from {\em networking}, smartphones and XR devices connect to WiFi and 3G/4G/5G networks, with download speeds typically surpassing upload speeds. For example, the Snapdragon X75 5G modem offers a downlink of up to 10 Gbps and an uplink of up to 3.5 Gbps. Embedded devices and microcontrollers also provide Ethernet connectivity, particularly in static deployment scenarios.

\subsection{System and Engine Layer} 

Above the hardware layer, resources are managed by operating systems and deep learning engines to facilitate the efficient runtime execution of models. Deep learning engines, in particular, are designed to operate across diverse operating systems and heterogeneous hardware. Additionally, scheduling the learning tasks on both cloud and mobile devices is crucial to enhance resource utilization and improve task collaboration. (1) From {\em operating systems}, cloud servers typically run on Linux or Windows, while mobile devices are more fragmented. Smartphones operate on Android, iOS, or HarmonyOS. XR devices use specialized operating systems like Apple's VisionOS and Meta's Horizon OS. Embedded devices and microcontrollers often run on lightweight Linux distributions or real-time operating systems (RTOS) that are event-driven and preemptive, or they may even function without any operating system at all, allowing for greater flexibility in manual resource scheduling; and (2) from {\em deep learning engines}, mainstream general-purpose options for cloud servers include TensorFlow \cite{proc:osdi16:tensorflow} and PyTorch \cite{proc:nips19:pytorch}. For mobile devices, mainstream engines include TensorFlow Lite or LiteRT \cite{link:tf:lite,proc:mlsys21:tf:lite:micro}, PyTorch Mobile \cite{link:pytroch:mobile}, TVM \cite{proc:osdi18:chen}, and MNN \cite{proc:osdi22:walle}. In addition to general-purpose engines designed for conventional deep neural networks, Transformer-based large generative models are driving the development of specific libraries and distributed training frameworks on the cloud, such as Hugging Face Transformers \cite{link:huggingface:transformers}, DeepSpeed \cite{link:deepspeed}, and Megatron-LM \cite{link:Megatron-LM}; as well as on-device deployment solutions, such as llama.cpp \cite{link:llama:cpp}, MNN-LLM \cite{proc:osdi22:walle}, MLC-LLM \cite{link:mlc-llm}, MediaPipe LLM \cite{link:mediapipe:llm}, and ExecuTorch \cite{link:executorch}.

\subsection{Model and Algorithm Layer}

Based on the deep learning engine, diverse models can be executed with different learning algorithms on the cloud and mobile devices to process various data types in practical applications. Oriented by learning tasks with inference and/or training phase, different tasks or split subtasks within a given task are allocated between the cloud and the mobile devices. The collaborative learning algorithm framework allows the use of cloud-based large model in conjunction with on-device small model to effectively and efficiently complete the tasks. The key objective is for the cloud-based large model to achieve strong performance across global data from all mobile devices, while the on-device small model optimizes for the specific user's local data. 

When selecting models and algorithms for cloud and mobile devices, the first consideration should be the specific data processing requirements of the learning task. (1) For {\em images and videos}, fundamental learning tasks in the fields of computer vision and computer graphics encompass detection, classification, segmentation, generation, and reconstruction. Prominent model architectures for these tasks include deep neural network (DNN), convolutional neural network (CNN), residual network (ResNet), generative adversarial network (GAN), and stable diffusion; (2) for {\em text and audio}, basic tasks in the field of natural language processing (NLP) include next-word prediction, machine translation, text summarization, text classification, sentiment analysis, named entity recognition (NER), speech recognition, and text-to-speech (TTS). Common model architectures include Transformer, long short-term memory (LSTM) network, recurrent neural network (RNN), and CNN; (3) for {\em other types of time-series data}, such as user behavior data and sensor data, the primary task remains sequence modeling. These data types can still adopt network architectures similar to those used in NLP; and (4) for {\em multimodal data}, in addition to representing each unimodal data source, crucial tasks include cross-modal alignment and fusion, typically using contrastive learning, canonical correlation analysis, domain adaptation, latent space modeling, or Transformer-based cross sequence modeling.

Besides the practical requirements for effective data processing, the selection of models and learning algorithms must also consider the hardware resources of cloud servers and mobile devices, as detailed in Section \ref{sec:framework:hardware}, along with the deployment costs. For example, cloud servers, which are resource rich, can support model parameters scaling up to trillions. To serve a large number of users cost-effectively, the model should be scaled down. In contrast, mobile devices, which are resource constrained, typically restrict model parameters to under billions. If on-device training is required, the model should be smaller. Models are also released in various parameter sizes to meet diverse application requirements. For conventional networks, options include the VGG and ResNet series. For large generative models, Llama 3.2 offers large language model (LLM) options with parameter sizes of 1B and 3B for on-device use, and multimodal options with parameter sizes of 11B and 90B, while Qwen 2.5 provides more LLM options with parameter sizes of 0.5B, 1.5B, 3B, 7B, 14B, 32B, and 72B, and vision-language options with parameter sizes of 3B, 7B, and 72B.

\subsection{Application Layer} 

At the top application layer, cloud-based large model and on-device small model are deployed to jointly serve real-world scenarios. (1) For {\em computer vision and computer graphics scenarios}, typical applications include facial recognition, video analysis in urban surveillance, object detection and tracking, scene segmentation in autonomous driving, image and video generation, and 3D object and environment interaction on smartphones and XR devices; (2) for {\em natural language scenarios}, typical applications include intelligent keyboard (e.g., Google's Gboard, Tencent's Wechat keyboard), virtual assistant (e.g., Apple's Siri, Amazon's Alexa, Huawei's Xiaoyi, Xiaomi's Xiaoai, OPPO's Xiaobu), and chatbot (e.g., OpenAI's ChatGPT, Alibaba's Qwen, Anthropic's Claude, Google's Gemini); (3) for {\em user behavior data analysis scenarios}, common applications include recommender systems for e-commerce platforms (e.g., Amazon, Taobao, Jingdong, Meituan), social media networks (e.g., Facebook, X, Weibo), and online video services (e.g., YouTube, TikTok, Kuaishou); (4) for {\em sensor data analysis scenarios}, typical applications include environmental monitoring, healthcare and fitness tracking, smart home automation, industrial machinery maintenance, manufacturing and quality control, and transportation systems optimization; and (5) for {\em multimodal data analysis scenarios}, typical applications including multimedia content understanding, visual question answering (VQA), interactive virtual assistants, multimodal content generation, autonomous systems and robotics, and healthcare and medical imaging.

\begin{figure}[!t]
  \centering
  \includegraphics[width=0.95\columnwidth]{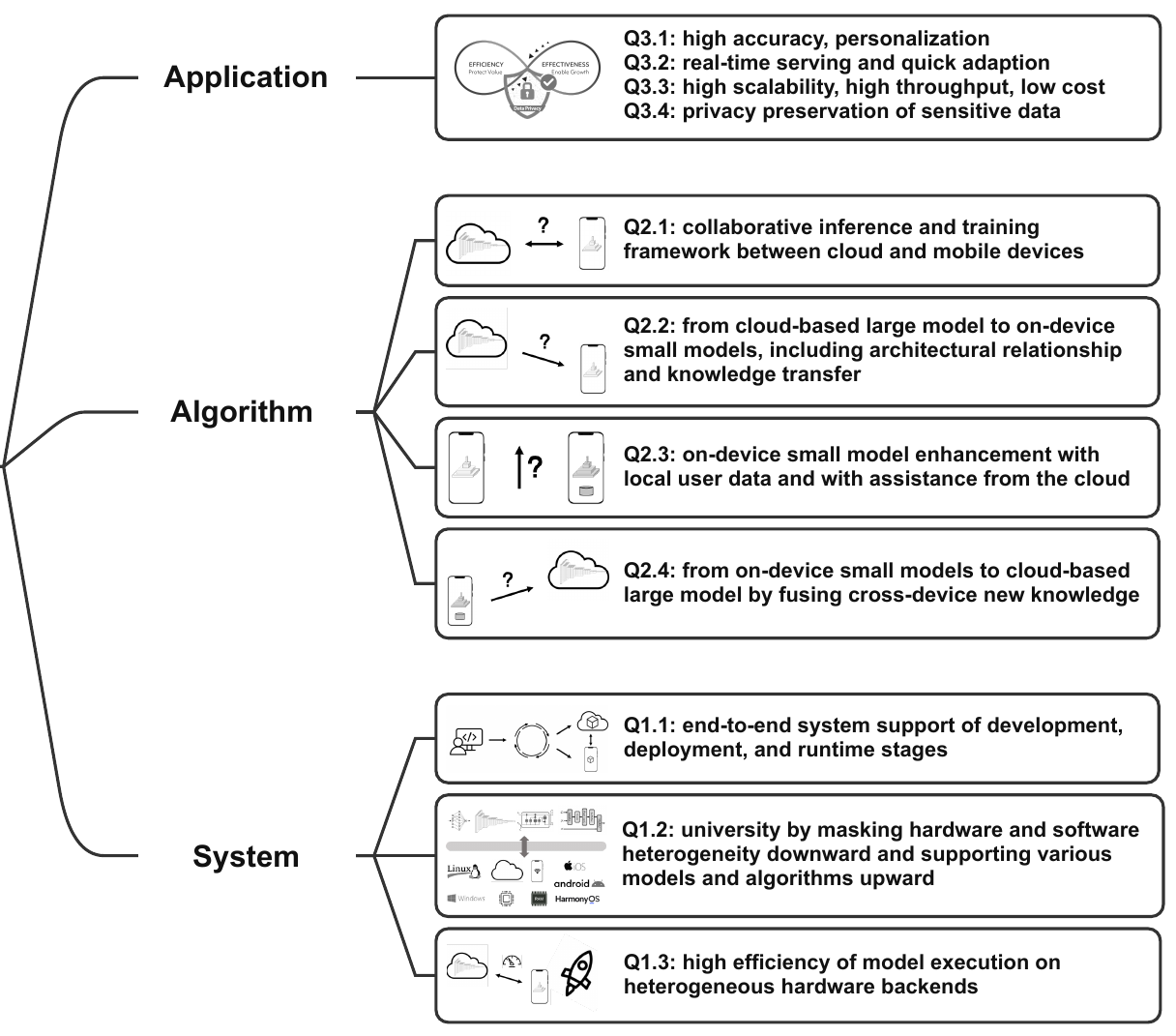}
  \caption{Key problems at different layers.}\label{fig:problem}
  \Description{See the caption.}
\end{figure}

\section{Key Problems At Different Layers}\label{sec:key:problem}

Given the existing hardware configurations of cloud servers and mobile devices, we list the key problems at different layers to facilitate collaborative learning of cloud-based large model and on-device small model.   

\subsection{Learning System and Engine Problems}

The first problem is {\em how to build an end-to-end system pipeline, supporting the development, deployment, and runtime stages of on-device small model and cloud-based large model collaborative learning.} The key challenges include the multi-granularity model deployment requirements on a vast number of devices with application-level, user-level, and device-level differentiation; the timeliness and the robustness guarantees of deploying models on mobile devices with dynamic execution environments; the frequent experimentation and fast iteration requirement for verifying the effectiveness of different models and learning algorithms; as well as the model training and inference runtime support and the low-latency data transfer pipeline between cloud and mobile devices for collaborative learning. In particular, different from cloud servers with stable wired connections, mobile devices with unstable wireless connections are intermediate available. Smartphones also allow only one mobile application (APP) to run in the foreground, while users tend to switch mobile APPs frequently. As a result, from the perspective of a certain mobile APP, each mobile device's availability is dynamic. Additionally, a mobile APP runs as a single process. The failure of any model execution task will lead to the crash of the whole APP, seriously impacting user experience.

The second problem is {\em how to ensure the universality of the learning system and engine, involving the downward cross-platform requirement of masking the heterogeneity of hardware and operating systems across cloud servers and mobile devices, and the upward capability of supporting a wide range of model architectures and learning algorithms.} Specifically, cloud servers and mobile devices differ significantly in hardware, such as processor, instruction set architecture, and memory configuration, as well as in operating system. Among mobile devices, the hardware and software ecosystems are even more fragmented. Moreover, the combination of resource heterogeneity and diverse learning task requirements necessitates varied network architectures. Consequently, supporting the execution of heterogeneous models across heterogeneous hardware backends presents a significant challenge.

The third problem is {\em how to guarantee the high efficiency of model execution on both cloud servers and mobile devices.} Beyond universality, the learning system and engine must deliver exceptional performance. This necessitates optimization at both the tensor operator and compute graph levels, tailored to the specific characteristics of heterogeneous hardware backends and various model architectures. Additionally, given the resource constraints and the dynamically changing resource status of mobile devices, it is challenging to schedule on-device hardware resources and determine the trigger time of cloud collaboration, thereby maximizing resource utilization and minimizing power consumption, all while maintaining low latency in task completion.

\subsection{Model and Learning Algorithm Problems} 

The first problem is {\em how cloud-based large model and on-device small model can effectively and efficiently collaborate in inference and/or training}. This further involves the following key issues: (1) how to design the interaction paradigm, particularly, whether it be single-device-to-cloud interaction or multi-device-to-cloud interaction; (2) how to allocate inference or training subtasks to the cloud and mobile devices; (3) what and when to exchange between the cloud and the mobile devices to facilitate learning task collaboration, such as exchanging data, feature, model parameter (update), and/or inference result; and (4) how to guarantee collaborative learning effectiveness, such as model convergency, generalization capability, and inference accuracy, given the misalignment settings of global and local optimization objectives, global and cross-device local data distributions, and large and small model architectures.

The second problem is {\em from the cloud downward to mobile devices, how to obtain on-device small model given the existence of cloud-based large model, and how to transfer knowledge from large model to small model.} The core issue lies in identifying the architectural relationship between on-device small model and cloud-based large model, which must allow the small model to meet the learning task requirements and adhere to the resource constraints of mobile devices, while supporting the collaborative learning framework and enabling bidirectional knowledge transfer. The small model could be a split part of the large model, a compressed lightweight version, or employ an entirely different architecture. Whether the small models on different mobile devices should take the same architecture or different architectures is still an open problem. The knowledge transfer method should also be designed in correspondence to the model architecture.  

The third problem is {\em on the side of mobile devices, how to enhance the performance of small model with local user data and with assistance from the cloud.} If on-device training of the small model is feasible, the key challenges lie in the scarcity of local user data, leading to the dilemma of few-shot learning with high generalization error; the discrepancy between global and local distributions, causing optimization bias in model finetuning; and the lack of labeled samples, particularly for computer vision tasks. If full model fine-tuning is infeasible on a mobile device (e.g., for billion-scale LLMs) due to memory constraints, one way is to consider memory-efficient tuning, and the other way is to consider enhancing small model during inference or test time. However, any test-time enhancement must maintain low latency to ensure that the real-time performance of on-device inference service is not compromised. 

The fourth problem is {\em from mobile devices upwards to the cloud, how to enhance the performance of cloud-based large model by fusing the new knowledge from on-device small model.} (1) Uploading samples for retraining the cloud-based large model is inefficient, not only wasting mobile device resources in local model enhancement but also potentially raising privacy concerns; (2) fusing new features from on-device personalized small models into cloud-based large model may also encounter feature space misalignment issues. This misalignment occurs because the feature space of each on-device small model, after local enhancement, is no longer consistent with that of the large model; and (3) aggregating parameter updates from small models across different mobile devices and applying the aggregate update to the cloud-based large model requires a strict architectural alignment between the large and small models. For example, it is challenging to aggregate the parameter updates from small models with different architectures, and it is also challenging to apply the aggregate update to the large model if the small models and the large model take totally different architectures. 

\subsection{Application Problems} 

The first problem is {\em from the research and development perspective, how to build practical, large-scale, and diverse datasets and corresponding performance metrics for evaluating the feasibility of collaborative learning algorithms and models}. (1) The practical requirement not only indicates that the samples should be collected from real-world scenarios but also implies that the datasets need to reflect real-world, imbalanced, and non-independent and identically distributions (non-iid) across different mobile devices. In particular, conventional datasets are suitable for evaluating cloud-based large models. However, for on-device small models, it is necessary to construct datasets that include user-level or device-level information to facilitate natural sample partitioning; (2) the large-scale requirement pertains not only to the size of the samples but also to the number of mobile devices or users involved, thereby validating the scalability of the proposed design; and (3) the diversity requirement encompasses a wide range of learning tasks, thereby validating the generality of the proposed design. Besides datasets, it is also crucial to reassess evaluation metrics, which involves two key issues: (1) defining a collaborative learning metric that encompasses both the performance of cloud-based large model and the personalized performance of on-device small models; and (2) integrating model performance with practical overhead, such as deployment cost, latency, resource consumption, and power consumption, which are especially significant for mobile devices. 

The second problem is {\em from the production perspective, how to meet the practical requirements of different applications and put the collaborative learning design of on-device small model and cloud-based large model into large-scale production use.} (1) For computer vision and computer graphic applications, key requirements include high service quality, low latency, high throughput, fast adaption to local dynamic data, and privacy preservation of sensitive images and regions; (2) for natural language applications, key requirements include real-time and accurate understanding and generation, personalization, high scalability and low cost for serving numerous users with long context, and privacy preservation of sensitive texts; (3) for recommender systems, key requirements include personalization, as well as real-time serving and adaption; (4) for sensor data analysis applications, key requirements include low latency, high accuracy, fast adaption to dynamic environments, and privacy preservation; and (5) for multimodal data analysis applications, key requirements include high accuracy, low latency, and high throughput.

\section{Current Advances}\label{sec:advance}
Specific to the listed key problems, we first review the advances at the system and the algorithm layers, then introduce existing datasets, evaluation metrics, and benchmarks, and finally present typical industrial applications.

\subsection{Advances At System and Engine Layer}

\subsubsection{Learning Engine} 
From the development of universal and highly efficient engines to support model execution on cloud server and mobile devices, deep learning frameworks (e.g., TensorFlow \cite{proc:osdi16:tensorflow} and its Lite/Micro version \cite{link:tf:lite,proc:mlsys21:tf:lite:micro}, PyTorch \cite{proc:nips19:pytorch} and its Mobile version \cite{link:pytroch:mobile}) normally require human experts to do operator-level and computation graph-level optimizations for heterogeneous hardware backends. With the explosion of models and frameworks at the top and the explosion of hardware at bottom, the workload of manual optimization is quite heavy that only some common cases can be covered. \citet{proc:osdi18:chen} built an automated optimizing compiler TVM, which takes a high-level specification of a deep learning model from existing frameworks and generates low-level optimized code for a diverse set of hardware backends, including lower-power CPUs, mobile GPUs, server-class GPUs, and FPGA accelerators. To combine the high performance of manual optimization and the low cost of auto optimization, \citet{proc:osdi22:walle} proposed a semi-auto search mechanism, which not only conducts manual operator-level optimization for heterogeneous backends but also facilitates identifying the best hardware backend with minimum cost for a computation graph, through searching the optimal implementation algorithm with optimal parameter settings for operators on available backends. Specific to large autoregressive generative models, existing learning engines have undergone significant adaptation, introducing specialized frameworks, such as MediaPipe LLM \cite{link:mediapipe:llm}, torchchat \cite{link:pytroch:torchchat} and ExecuTorch \cite{link:executorch}, MLC-LLM \cite{link:mlc-llm}, and MNN-LLM \cite{proc:osdi22:walle}. These frameworks mainly focus on accelerating matrix multiplication operations during the prefill phase to reduce computational overhead, optimizing the layout and management of key-value (KV) caches in the decoding phase to improve memory efficiency, and ensuring compatibility with advanced quantization algorithms for efficient deployment without sacrificing model performance.

\subsubsection{System Platform and Scheduling Mechanism}
From the end-to-end system support of development, deployment, and runtime stages,  \citet{proc:sysml19:google:fl} focused on cross-device federated learning framework and enabled engineers to develop machine learning tasks with different configurations in Python and TensorFlow, conduct simulative debugging, and finally deploy the tasks in production environment. They proposed round-level protocol for collaborative learning between mobile devices and cloud with the key phases of mobile device selection on cloud, cloud-to-mobile device model and configuration synchronization, mobile device-to-cloud model update reporting, and server-based model update aggregation. They also designed the system modules of local data management, programmatic configuration, job scheduling and  invocation, task execution, and reporting for mobile device, as well as a hierarchical system architecture for server around actor programming model with the actors of coordinators, selectors, master aggregators, and aggregators. \citet{proc:osdi22:walle} did not restrict the collaborative learning framework between mobile devices and cloud, and built a general-purpose and large-scale production system, called Walle. Walle consists of a deployment platform, a data pipeline, and a compute container, supporting the development, deployment, and runtime stages of a machine learning task. In particular, the compute container is based on the deep learning engine MNN and exposed to a Python thread-level virtual machine, meeting the practical industrial need of frequent experimentation and deployment for daily task iteration. In addition to industrial end-to-end platforms, some other existing work focused on how to facilitate the collaboration between mobile devices and cloud at the system level. \citet{proc:mobisys10:MAUI} considered how to offload mobile code to cloud, leveraged managed code environments for dynamic offloading without requiring programmer intervention or whole-process migration, and dictated how to partition the application at runtime to maximize energy savings under the current networking conditions. They also showed case applications of face recognition, voice-based language translation, and arcade game. \citet{proc:osdi12:comet} designed a distributed shared memory method to reduce communication for code offloading between mobile device and cloud, allow computation to resume on the mobile device if the server is lost at any point during execution, and also support multi-threaded applications. \citet{proc:eurosys11:clonecloud} focused on general mobile applications and built CloneCloud to offer flexible application partitioning and execution, allowing unmodified mobile applications to offload parts of their processing to cloud-based clones. CloneCloud migrates application threads from mobile devices to cloud clones during execution, running them there and then reintegrating them back. \citet{proc:mobisys15:replicate} proposed to replicate
mobile application on both mobile device and the server. Since execution on either side may be faster during different phases, their design allowed either replica to lead the execution. Hu et al. \cite{proc:sec19:linkshare:hu} studied concurrent and continuous interactions between mobile devices and cloud involving different server backends and proposed a mobile device-centric scheduling mechanism, called LinkShare, which extends the operating system scheduler for concurrent inter-application network-bound requests by incorporating earliest-deadline first with limited sharing strategy. Specific to computer vision applications and CNN models, \citet{proc:mobisys16:MCDNN} designed a scheduling mechanism to adaptively select model variants of differing accuracy and model execution platform (i.e., either mobile device or cloud) to keep within per-request resource constraints (e.g., memory) and long-term constraints (e.g., energy) while maximizing average classification accuracy.

\subsection{Advances At Model and Algorithm Layer}

\subsubsection{Task Splitting Strategy and Collaboration Framework Between Mobile Devices and Cloud}

Under the conventional cloud-based learning framework, the task splitting and collaboration strategy is that each mobile device uploads raw data, and the cloud trains a model on the aggregated global dataset, while during the real-time serving phase, mobile devices send queries to the cloud, and the cloud performs model inference and returns the prediction results. The primary workload is concentrated on the cloud side. To balance the workload on the sides of cloud and mobile devices, existing work studied the design of task splitting strategy for training and/or inference phase. 

For collaborative training between mobile devices and cloud, the cross-device federated learning framework proposed by \citet{proc:aistats17:fl} originally let the cloud and each mobile device hold the same small model that is deployable on resource-constrained devices and distributed both model training and inference tasks from the cloud to mobile devices, such that private user data will not leave local devices. \citet{proc:jnca18:fl:split} combined federated learning with splitting learning, which splits the low-level encoder to the mobile device and keeps the other layers on the cloud for collaborative training. \citet{proc:icml21:fl:split:per:head} considered collaborative training of encoder for shared representation across different mobile devices and on-device local training of personalized header. \citet{proc:mobicom20:submodel} proposed a federated submodel learning framework, allowing the cloud to keep a full large model, while each mobile device just maintains a small part of the large model (e.g., the embedding vectors of one user's local few items plus the other network layers for a recommendation model), called submodel. In particular, the personalized submodel is independently executable on each mobile device. \citet{proc:mobicom21:li:hermes} focused on computer vision models and proposed to apply structured pruning to prepare a small subnetwork for each mobile device, while the cloud server aggregates only the overlapped parameters across different subnetworks. \citet{proc:nips22:FedRolex} proposed to extract a small submodel for each mobile device from the large global model on the cloud using a rolling window that advances in each communication round. Since the window is rolling, submodels from different parts of the large global model are extracted in sequence in different rounds, and all the parameters of the global model are evenly trained over the local data of mobile devices. \citet{arxiv23:dccl} regarded the ensemble of an on-device small model and a cloud-based large model as a global serving model, where the small model and the large model can take totally different network architectures, such as CNN and Transformer. The proposed design allowed each mobile device and the cloud to jointly finetune the serving model over new data in a parameter-efficient way (i.e., freezing the large model) with the help of a distilled version of the large model. \citet{arxiv23:offsite:tuning} proposed an offsite-tuning framework, which compresses the large middle layers of a model into a light-weight emulator using layer-drop and facilitates downstream users with resource-constrained devices to finetune the top and bottom layers without accessing the full model. \citet{proc:sec2019:collaborative} studied the user behavior prediction task with RNN and leveraged knowledge distillation to mutually and continuously transfer the knowledge between on-device small models and cloud-based large model, thereby mitigating data heterogeneity and data drift over time. In particular, the large model and the small model have different
sizes of LSTM units and different embedding sizes. \citet{proc:kdd21:patch} targeted on the recommendation model, relied on patch learning for on-device model personalization, and adopted model distillation to integrate the patches from mobile devices into the cloud-based global model. \citet{proc:iclr24:llmfinetune:emulator} and \citet{proc:colm24:proxytuning} focused on LLM and proposed an emulated or proxy fine-tuning framework to enable pretraining and finetuning at different model scales (e.g., a large 70B model for pretraining and a small 7B model for finetuning). A special case, called LM up-scaling, avoids resource-intensive fine-tuning of large pre-trained models by ensembling them with small fine-tuned models, essentially emulating the result of fine-tuning the large pre-trained model.

For collaborative inference between mobile devices and the cloud, early work considered how to filter data on each mobile device before sending them to the cloud. \citet{proc:mlsys19:video:edge:classifier} and \citet{proc:sigcomm20:video:diff} proposed to do frame filtering with on-camera model to facilitate cloud-based visual analytics. \citet{proc:sigcomm20:resend} adopted an interactive workflow, where cameras initially upload low-quality video streams and re-transmit high-quality video of several key regions based on feedback from the cloud, improving inference accuracy. \citet{proc:icdm23:livestream} considered mobile livestream applications in e-commerce and proposed a two-stage product recognition framework. Initially, an on-device unimodal (i.e., visual) model processes the frames. Only for a small proportion of frames, where the on-device model exhibits low confidence, their extracted unimodal features are then forwarded to the cloud for multimodal understanding. \citet{proc:kdd25:lv:llm:rec} focused on recommender systems and proposed to leverage LLM for cloud-based item matching and conventional light-weight recommendation model for on-device re-ranking. \citet{proc:acl24:query:large:small} proposed to leverage on-device small LLM to process privacy information (e.g., user profile) before sending to cloud-based LLM. They designed sketch-based and token probability-based collaborative generation methods. In the sketch-based design, on-device small LLM first generates a simplified sketch based on user query. This sketch retains key contextual information while omitting sensitive data, which is then transmitted to the cloud-based LLM to complete the generation task. The token probability-based approach enhances privacy protection by transmitting only the predicted results of the first few tokens instead of the entire response. At the level of whole user queries, existing work focused on routing user queries to either an on-device small model or a cloud-based large model, especially for LLMs. The major optimization objective is to reduce cloud service costs and response latency while ensuring overall response quality. Representative work include selective query \cite{proc:iclr13:class:router} for image classification models; AutoXPCR \cite{proc:kdd24:AutoXPCR} for time-series forecasting models; model multiplexing \cite{proc:nips23:llmselect}, hybrid LLM \cite{proc:iclr24:llm:route:hybrid}, RouteLLM \cite{proc:iclr25:llmroute}, EmbedLLM \cite{proc:iclr25:embedllm}, RouterDC \cite{proc:nips24:routerdc}, and ZOOTER \cite{proc:naacl24:llm:ensemble} for LLMs; and RouteT2I \cite{arxiv24:RouteT2I} for text-to-image generation models. Similar to model routing, another line of work on model cascade considered first invoking the on-device small model and then determining whether to call the cloud-based large model or not, such as FrugalGPT \cite{arxiv23:FrugalGPT}, LLM cascades \cite{proc:iclr24:llm:cascades}, AutoMix \cite{proc:nips24:automix}, Tabi \cite{proc:eursys23:llm:discriminative:route}, and adaptive teacher-student collaboration \cite{proc:cvpr24:student:teacher:colla}. Rather than in a cascade structure, some studies explored how small and large models can collaborate in a parallel way. \citet{proc:emnlp23:spec:decoding}, \citet{proc:icml23:spec:decoding}, and \citet{arxiv23:spec:decoding} proposed a speculative decoding framework for accelerating LLM inference, which leverages a small LLM to quickly and serially generate multiple tokens as a draft, and then lets a large LLM verify these tokens in parallel. \citet{proc:acl23:contrast:decode} designed a contrastive decoding framework to improve LLM generation quality by exploiting the contrasts between large and small LLMs through choosing tokens that maximize their log-likelihood difference, thereby producing high-quality text that amplifies the good behavior of the large LLM and diminishes the undesired behavior of the small LLM. \citet{arxiv:fast:slow} studied how small and large LLMs work together during the decoding process. They proposed to unify speculative decoding, contrastive decoding, and emulated or proxy tuning by framing them through a fast and slow generating framework, where the large model is characterized as System 2, which operates slowly and deliberately, while the smaller model is System 1, functioning quickly and intuitively. Different from selecting or combining small and large models to answer different queries, some existing work considered model splitting between mobile devices and cloud. \citet{proc:asplos17:dnn:split} proposed to horizontally partition DNN between a mobile device and the cloud at the granularity of network layers (e.g., the mobile device executes the initial layers of inference, while the cloud handles the remaining layers.) to reduce latency and mobile energy consumption. \citet{proc:mobicom22:huang:agilenn} proposed to vertically split the outputs of the low-level feature extractor, and let the cloud and the mobile device separately complete the remaining forward processes, the outputs of which are combined as the inference results. In addition to model splitting, \citet{proc:icdcs17:early:exit} and \citet{proc:mobicom20:early:exit} introduced early-exiting strategies \cite{proc:icpr16:early:exit}, enabling inference results to be flexibly output at intermediate layers on either the mobile device or the cloud, rather than being constrained to the final exit on the cloud. \citet{proc:nips22:llm:early:exit} designed a token-wise early-exiting strategy for LLMs with with per-token confidence measure at different Transformer layers. 

According to what the on-device small model and the cloud-based large model exchange, existing work mainly can be categorized into {\em data-based (e.g., exchanging raw samples or user queries), feature-based (e.g., exchanging intermediate or final outputs of models), and parameter-based (e.g., model or model update)}. In particular, typical {\em data-based} collaboration mechanisms include conventional cloud-based training and inference framework, data filtering, and query routing; typical {\em feature-based} collaboration mechanisms include model splitting, early exiting, distillation, and parallel decoding; and typical {\em parameter-based} collaboration mechanisms include federated learning and its variants, model ensemble, and offsite and emulator or proxy tuning.

\subsubsection{From Cloud-Based Large Model to On-Device Small Model}

To meet on-device task requirement and deployment constraints, light-weight models can be manually designed from scratch or automatically constructed using neural architecture search (NAS) \cite{proc:iclr17:nas,arxiv:mobilenetv1,proc:cvpr18:mobilenetv2,proc:iccv19:mobilenetv3,proc:cvpr19:mnasNet,proc:icml19:efficientnet}. In addition, on-device model can also be compressed from the cloud-based large model using pruning \cite{proc:nips15:pruning,proc:iclr19:lottery} and quantization \cite{proc:icml15:quantization,proc:iclr16:deep:compress} algorithms. Specifically, {\em pruning} intends to remove redundant parts of a large model, such as weights, neurons, or layers, that do not significantly contribute to the model's predictive performance. According to whether the changes in the architecture of the network are regular or not, pruning can be further categorized into structured pruning \cite{proc:iccv17:channel:pruning,proc:nips23:llm:pruner,proc:iclr25:llm:struc:pruning} and unstructured pruning \cite{proc:nips15:pruning,proc:icml:sparsegpt}. The resulting model sparsity can be exploited by hardware and software for computation acceleration and memory efficiency \cite{arxiv18:sparsity:acc}. {\em Quantization} intends reduces the precision of model weights and activations from floating-point (e.g., FP32) to lower-bit representations (e.g., INT8 and INT4) to improve memory and computation efficiency. Based on whether quantization is integrated with training, it can be categorized into: quantization-aware training (QAT) \cite{proc:cvpr18:quantization:aware:training, proc:acl24:qat:llm}, which simulates quantization effects during training, allowing the model to adapt and maintain higher accuracy; and post-training quantization (PTQ) \cite{proc:icml15:quantization,proc:iclr16:deep:compress,proc:eccv16:2bit:quant,arxiv22:gptq,proc:nips22:ptq,proc:icml23:ptq:llm,proc:mlsys24:awq}, which applies quantization to a pre-trained model without retraining, using calibration techniques to reduce accuracy loss. In addition, based on whether the same or different bit-widths are used across different parts of the model, quantization can be categorized into uniform quantization, which applies a fixed bit-width to all parameters (e.g., weights and activations) throughout the model; and mixed precision quantization \cite{arxiv16:DoReFa-Net:mix:quat,proc:cvpr19:mixed:quat,proc:iccv19:mix:quat,proc:nips22:gpt3int8}, which assigns varying bit-widths to different layers, or even different parameters within a layer, based on their sensitivity to quantization, for better accuracy and efficiency trade-off.

To transfer knowledge from cloud-based large model to on-device small model, knowledge distillation (KD) \cite{arxiv:knowledge:distill} is a widely used technique. The small model as a student learns to imitate the behavior of the large model as a teacher. In standard KD \cite{arxiv:knowledge:distill}, the small student model learns from the large teacher model's logits, which serve as soft labels to guide training. The follow-up work introduced different forms of supervised training signals from the teacher model, including response-based, feature-based, and relation-based knowledge. For response-based distillation, \citet{dml} extended the conventional teacher-student paradigm by allowing an ensemble of students to learn from each other’s outputs; \citet{noisystudent} added noise into the student’s training process when distilling teacher's output, improving generalization and robustness compared to the large teacher model. For feature-based distillation, which leverages intermediate representations, \citet{fitnets} explored a student model that is deeper and thinner than the teacher. They proposed to utilize not only the teacher model's outputs but also its intermediate representations as hints to enhance the student model's training process and final performance. Because the student model's intermediate hidden layer is generally smaller than the teacher model’s intermediate hidden layer, additional parameters were introduced to map the student hidden layer to the prediction of the teacher hidden layer. For relation-based distillation, which focuses on inter-layer relationships, \citet{relation} proposed to transfer the flow between layers to the student by computing the inner product of features from two layers, mirroring problem-solving processes.

Specific to LLMs, existing work can be categorized into white-box and black-box distillation based on the accessibility of the large teacher LLM's output distribution. In {\em white-box distillation}, the output distribution of the open-source teacher LLM is accessible, allowing the small student LLM to closely match it. However, unlike traditional tasks such as image classification, the output space in text generation, defined by the tokenizer's vocabulary, is significantly more complex, making conventional forward Kullback-Leibler (KL) divergence prone to mode-averaging issues in predicting next token. \citet{minillm} thus proposed MiniLLM to minimize reverse KL divergence, allowing the small student LLM to seek the major modes of the teacher LLM's outputs. \citet{distillm} further designed DistiLLM using skew KL and reverse KL divergence loss to guarantee stable gradients and small approximation error. They also introduced an adaptive off-policy strategy to effectively utilize the generated outputs of the student LLM. \citet{revisitkd} decomposed KL divergence into a target-oriented and a diversity-oriented component, applying adaptive teaching strategies for easy-to-learn and hard-to-learn tokens, respectively. To mitigate the train-inference distribution mismatch, \citet{gkd} designed on-policy KD by incorporating student-generated outputs into the distillation process, such that the small student LLM can receive token-specific feedback from the teacher's logits. \citet{ddk} focused on domains with huge performance gaps when distillation, assigning them higher sampling weights to maximize distillation benefit. From a reinforcement learning perspective, an autoregressive generation model can be interpreted as a language policy, where the model predicts the next token (i.e., action) based on the current context (i.e., state). Within this framework, KD corresponds to behavior cloning in imitation learning \cite{seqkd, imitation}. \citet{momentmatching} formulated KD in the context of imitation learning, derived the imitation gap between the large teacher LLM and the small student LLM via moment-matching distance, and proposed an adversarial training algorithm based on this gap. In {\em black-box distillation}, where the teacher model is closed-source, the student model can't get access to the internal information of the teacher, such as logits and hidden states. A straightforward way of distillation is to pretrain or fine-tune the small student LLM with teacher-generated outputs \cite{seqkd, alpaca, vicuna}. For instruction-following tasks, \citet{wizardlm} proposed to collect large-scale instruction data with varying levels of complexity from the large teacher LLM to enhance the small student model’s instruction-following capabilities. \citet{distillstep} and \citet{cotdistill} explored chain-of-thought (CoT) distillation, providing richer training signals to transfer black-box teacher LLM's reasoning ability to the small student LLM. \citet{metamath} and \citet{orcamath} focused on mathematics, expanding seed datasets using rule-based data augmentation. More recently, DeepSeek-R1 \cite{deepseek}, trained via reinforcement learning, has been leveraged as a large teacher model to distill multiple small models through supervised fine-tuning on its generated reasoning data. Notably, it highlights that small student models struggled to develop reasoning abilities through self-reinforcement learning due to capacity limitations.

\subsubsection{On-Device Small Model Enhancement With the Coordination of Cloud}
In addition to the fundamental requirements for lightweight deployment and the knowledge transfer process from the cloud for effective initialization, on-device small models can be further enhanced using local user data through finetuning, as well as incremental learning and transfer learning techniques for conventional neural networks. To reduce the memory overhead of on-device training, where the activations dominate training memory \cite{arxiv19:memory,proc:nips20:tinytl}, the strategies of setting a small training batch size, layer or parameter freezing, inserting a light-weight adapter \cite{proc:icml19:adapter}, low-rank adaption (LoRA) \cite{proc:iclr22:lora}, prefix or prompt tuning \cite{proc:acl21:prefix:tuning,proc:emnlp21:prompt:tuning}, low-precision representation \cite{proc:nips23:qlora}, and recomputation \cite{proc:mobisys22:xu:memory} were proposed. Alternatively, memory-efficient retrieval augmentation methods (e.g., \cite{proc:nips20:rag, proc:icml20:rag, proc:iclr20:knnlm}) can be applied to LLMs. Search-R1 \cite{arxiv25:searchR1} also demonstrated the potential of small LLMs to reason and utilize external tools, particularly search engines, through reinforcement learning. 

However, the data available on each individual mobile device are often limited in size and of lower quality compared with cloud-based global data pool, which causes on-device model enhancement to face the overfitting challenges associated with few-shot learning. Collaborating with the cloud, or with other mobile devices under the cloud’s coordination, is a natural solution. Under the {\em data-based collaborative learning} framework between each mobile device and the cloud, \citet{proc:kdd22:mpda}, \citet{arxiv22:coda}, and \citet{proc:mobicom24:gong} proposed to retrieve samples from the cloud pool to augment on-device local dataset with domain adaptation techniques, two-class classifier between local and global data, and data matching strategies, respectively. Specific to LLMs, \citet{proc:kdd24:ding} explored storing each user's historical interactions with a cloud-based LLM on the mobile device as a local datastore to enhance the generation of an on-device small LLM using retrieval augmentation. Since retrieval augmentation occurs during real-time generation, they also proposed a subset selection mechanism to maintain the local datastore within a certain size threshold, ensuring it meets real-time generation requirements. Under the {\em parameter-based federated learning} framework \cite{proc:aistats17:fl}, besides the original objective of global model optimization, \citet{arxiv20:fl:per} proposed three general frameworks to learn personalized models for different mobile devices, including user clustering, model interpolation, and data interpolation. Some other work \cite{proc:nips17:fl:mtl,arxiv19:fl:meta,proc:nips20:pfl:meta,proc:iclr21:mul:model,proc:nips22:fl:contrast,proc:cvpr22:pfl} also studied how to enable personalized federated learning through multi-task learning, meta learning, contrastive learning, or fine-grained model aggregation. Under the {\em feature-based collaborative learning} framework, \citet{proc:aaai22:fedproto} proposed to let the cloud and mobile devices exchange class prototypes, namely, feature vectors representing each class. The cloud aggregates local prototypes from multiple mobile devices and then distributes the global prototypes back to all devices, thereby regularizing the training of local models. \citet{proc:cvpr23:fed:proto} further introduced contrastive learning to mitigate the domain heterogeneity across different mobile devices, by pulling the sample feature closer to cluster prototypes belonging to the same semantics than cluster prototypes from distinct classes, while aligning the local features with the respective unbiased prototype. To avoid model training on resource-constrained mobile devices while still adapting to local data distributions, \citet{proc:www23:device:model:generation} proposed to let each mobile device upload its real-time features, encoded by a static encoder, to the cloud. The cloud then generates dynamic parameters corresponding to local features using hierarchical hypernetworks and sends them back to the mobile devices. The static encoder and the dynamically generated layers are stacked together for real-time on-device inference.

\subsubsection{Cloud-Based Large Model Enhancement With the Awareness of Mobile Devices}

After personalized enhancement of on-device small models, the cloud-based large model can also be enhanced by perceiving each mobile device user's personality or fusing the new knowledge of different mobile devices. Conventional cloud-based learning framework directly {\em collects raw samples} from mobile devices for training and inference. \citet{proc:kdd25:ding} focused on NLP scenarios where text data cannot be uploaded alongside explicit user identifiers, yet cloud-based large models still require personalized service. To reconcile the need for both anonymity and personalization, they proposed that each mobile device learns the parameters of a local distribution, which is used to dynamically sample user embeddings that can be inserted as the input prefix of cloud-based LLM. Instead of a static user embedding, the sampled user embedding, along with the text data, is uploaded to the cloud, enabling personalized model adaptation while preserving user privacy. Under the {\em parameter-based federated learning} framework \cite{proc:aistats17:fl}, each mobile device does not need to upload local private data to the cloud, and in each communication round, the cloud aggregates the model updates from participating mobile devices and updates the global model. The default aggregation algorithm is called federated averaging \cite{proc:aistats17:fl}. Some follow-up work considered more practical settings, such as cross-device data heterogeneity \cite{proc:iclr20:fedavg:noniid,proc:icml20:SCAFFOLD}, intermittent device availability \cite{jour:joc24:fedlaavg}, inconsistent on-device training iterations \cite{proc:mlsys20:fedprox,proc:nips20:fednova}, existence of stragglers \cite{proc:aistats21:dropout}, and different communication topologies \cite{proc:icml20:fl:topolgoy}, and proposed several variants of federated averaging to guarantee convergency. To aggregate the heterogeneous model updates from mobile devices that can differ in size, numerical precision, or structure, \citet{proc:nips20:fed:distill} proposed ensemble distillation for fusion by training the cloud-based large model through unlabeled data on the outputs of the heterogeneous models from mobile devices. Considering memory-efficient adaption methods may be applied for on-device training, some other work studied how to aggregate these light-weight adapters. \citet{proc:kdd21:patch} adopted model distillation to integrate adapters from mobile devices into cloud-based recommendation model. \citet{proc:cvpr24:orthogonal:LoRA} focused on using LoRA for text-to-image generation models and proposed to let each user samples a low-rank decomposed matrix for a specific concept from a shared orthogonal basis, such that synthesizing multiple user-specific concepts on the cloud does not lead to conflicts. \citet{proc:nips23:mix:concept} designed a cloud-based gradient/LoRA fusion algorithm for cross-device concept embedding tuning to mitigate multi-concept conflict. \citet{proc:emnlp24:heterLoRA} allowed heterogeneous LoRA ranks across mobile devices for their individual system resources, efficiently aggregated and distributed LoRA modules in a data-aware way by applying local rank pruning and sparsity-weighted aggregation on the cloud. Under the {\em feature-based collaborative learning} framework, \citet{proc:icml21:generator:agg} proposed to let the server ensemble the extracted features and the labels of local data on each mobile device using a conditional generator, which can also be sent back to each mobile device to facilitate local model training with knowledge distillation. \citet{proc:icdm23:livestream} considered the collaboration between on-device unimodal model and cloud-based multimodal model. The key problem is that after finetuning the unimodal model on local user data, the personalized unimodal features from mobile devices no longer align with the features in other modalities on the cloud. They thus designed a prompt generation module, transforming the unimodal feature from each individual mobile device as personalized prompt to guide cloud-based multimodal processing. \citet{proc:icml24:Burns} explored {\em weak-to-strong generalization}, where finetuning a large strong model with labels generated by a small weak model makes the large model consistently outperform the small model. They further introduced an auxiliary loss to reinforce the large model’s confidence in its own predictions, even when the large model disagree with the weak model's label. For example, when supervising GPT-4 with a GPT-2-level model on NLP tasks using the auxiliary confidence loss, nearly 80\% of the performance gap between the weak and strong models was recovered. \citet{arxiv24:llm:pretrain:slm} proposed to improve LLM pre-training efficiency and quality by leveraging a small LLM, which provides soft labels as additional training supervision and selects a small subset of informative and hard training examples, thereby effectively transferring the predictive distribution of the small LLM to the large strong LLM, while prioritizing specific regions of the training data distribution. \citet{proc:nips24:weak:to:strong:search} proposed to enhance a strong LLM with weak test-time greedy search by exploiting the log-probability difference between small tuned and untuned LLM as both reward and value to guide the decoding of the large model. \citet{arxiv24:multi:weak:to:strong} studied supervising a strong LLM with multiple small weak LLMs with different specializations under the framework of hierarchical mixture-of-experts.

\subsection{Existing Datasets, Metrics, and Benchmarks}

\begin{table}[!t]
\caption{Public datasets and metrics for collaborative learning of on-device small model and cloud-based large model.}\label{tab:benchmark}
\centering
\resizebox{\columnwidth}{!}{
\begin{tabular}{l|l|l|l|l}
\toprule
Task & Dataset & \# Total Samples & Partition By & Metric   \\ \midrule
\multirow{4}{*}{Image Classification} & FEMNIST \cite{arxiv18:dataset:leaf} &  805,263 & 3,550 Writers & Accuracy \\ \cline{2-5}
                                      & iNaturalist-User-120k \cite{proc:eccv20:fl:cv} & 120,300  & 9,275 Citizen Scientists &  Accuracy \\ \cline{2-5}
                                      & Landmarks-User-160k \cite{proc:eccv20:fl:cv} & 164,172 & 1,262 Authors & Accuracy\\ \cline{2-5}
                                      & CelebA  \cite{arxiv18:dataset:leaf}         & 200,288 & 9,343 Celebrities & Accuracy \\\midrule
\multirow{9}{*}{Natural Language}     & Sentiment140  \cite{arxiv18:dataset:leaf}    & 1,600,498 & 660,120 Twitter Users &  Accuracy \\  \cline{2-5}
        & Reddit \cite{arxiv18:dataset:leaf}    & 1,660,820  & 56,587,343 & Accuracy \\  \cline{2-5}
        & Shakespeare \cite{arxiv18:dataset:leaf} & 4,226,158 & 1,129 Roles &  Accuracy \\ \cline{2-5}
        & Amazon Kindle \cite{proc:kdd24:ding} & 25,600,000 &  5,600,000 Amazon Users & Accuracy \\\cline{2-5}
        & PersonalDialog \cite{arxiv19:PersonalDialog} & 20,830,000 & 8,470,000 Weibo Users & Accuracy \\ \cline{2-5}
        & Persona-Chat \cite{proc:acl18:personachat} & 10,981 & 1,155 Personas & PPL/F1 \\   \cline{2-5} 
        & LiveChat \cite{proc:acl23:livechat}   & 1,332,073 & 351 Streamers &  ROUGE/BLEU \\ \cline{2-5}
        & DuLeMon \cite{proc:acl22:baidu} & 27,501 & 1,687 Personas & PPL, BLEU \\ \cline{2-5} 
        & DailyDialog \cite{proc:ijcnlp17:DailyDialog}   & 13,118 & 1660 Users & PPL, BLEU\\
        \midrule
\multirow{3}{*}{Recommendation}     & MovieLens-1M  \cite{proc:nips22:submodel:opt}  & 1,000,209 & 6,040 Users &  Accuracy \\  \cline{2-5}
                 & Amazon Electronics \cite{proc:nips22:submodel:opt} &  123,147 &  1,870 Users & AUC \\ \cline{2-5} 
                 & Alimama \cite{link:alimama:dataset} & 26,000,000 & 1,140,000 Users & AUC\\\midrule
\multirow{4}{*}{Multimodal}   & CREMA-D  \cite{proc:kdd23:fl:multimodal}  & 4,798
    & 72 Actors &  UAR \\  \cline{2-5}
    & MELD \cite{proc:kdd23:fl:multimodal} & 9,718 & 86 Roles & UAR  \\ \cline{2-5}
    & KU-HAR \cite{proc:kdd23:fl:multimodal} & 10,300 & 66 Participants & F1 \\ \cline{2-5}
    & PTB-XL \cite{proc:kdd23:fl:multimodal} & 21,700 & 34 Patients & F1\\
\bottomrule
\end{tabular}
}
\end{table}

Different from conventional cloud-based learning that evaluates the model performance metrics on the whole dataset, collaborative learning of on-device small model and cloud-based large model not only needs to evaluate the large model's performance over the global data, but also needs to test the small model's performance on the serving user's local data. The data on different mobile devices are non-iid. Therefore, the unique requirement of datasets and metrics is to incorporate user-level or device-level information for natural dataset partition and performance evaluation. Existing benchmarks focused mainly on federated learning with both natural and synthetic data partitions for conventional discriminative tasks, such as LEAF \cite{arxiv18:dataset:leaf}, Federated Visual Classification \cite{proc:eccv20:fl:cv}, and FedMultimodal \cite{proc:kdd23:fl:multimodal}. For natural language generative task, there also exist some dialogues datasets  collected from real-world scenarios or scripts with natural user-level or role-level partition, such as PersonalDialog \cite{arxiv19:PersonalDialog}, Persona-Chat \cite{proc:acl18:personachat}, and LiveChat \cite{proc:acl23:livechat}. Table \ref{tab:benchmark} summarizes some representative datasets with natural user-level partition for different tasks as well as the corresponding evaluation metric. In particular, for {\em image classification task}, accuracy is the primary metric; for {\em NLP}, accuracy is used for {\em classification tasks} (e.g., sentiment analysis or next word/character prediction), while {\em response generation} is typically evaluated using metrics such as perplexity, ROUGE, or BLEU; for {\em recommendation task}, the area under the ROC curve (AUC) is the key metric; and for {\em multimodal task}, the F1 score and the unweighted average recall (UAR) (i.e., the sum of class-wise recall divided by number of classes) are commonly used. In addition to the original metrics for evaluating the global performance over the full test set, some variants were also defined for user-level personalized evaluation. In the LEAF benchmark, \citet{arxiv18:dataset:leaf} emphasized the importance of explicitly defining how accuracy is weighted across devices by considering whether each mobile device is weighted equally or each data sample is, which gives preferential treatment to power users or high-usage mobile devices. For example, for classification task, aggregated accuracy was computed as the test accuracy on each participating device weighted by its number of local samples \cite{proc:nips22:pfl:bench}; and for recommendation task, user‑level average AUC is proposed as the weighted average of per‑device AUCs, with each device’s AUC weighted by the size of its on‑device test set \cite{arxiv22:rec:model:label:correct,proc:kdd17:gauc}.

\subsection{Typical Industrial Applications}

Collaborative learning of on-device small model and cloud-based large model has been applied for large-scale production use in some typical industrial scenarios as follows.

\subsubsection{Cloud-Based Large Recommender System and On-Device Small Re-Rank Model Collaboration}

Traditional recommender systems were typically cloud dominated. When a user request is received from a mobile device, the system executes a pipeline consisting of matching, coarse ranking, and fine ranking stages, ultimately filtering and returning a shortlist of items, usually in the tens, to the mobile device. In particular, the fine ranking stage normally adopts high-capacity models, such deep neural networks for user behavior sequence modeling, to precisely score and sort candidate items. On the side of the mobile device, one page shows only a few items. In addition, as the user browses, his/her interests may shift dynamically, which, however, cannot be quickly perceived by the conventional cloud-based complex pipeline. As a result, those items that have not been exposed cannot be adjusted in low latency according to the user's evolving interests. Mobile Taobao, for the first time, introduced the new stage of on-device re-ranking \cite{proc:cikm20:edgerec} following the traditional cloud-based recommendation pipeline. Kuaishou also extended on-device re-ranking to short video recommendation scenarios \cite{proc:cikm22:kuaishou:re-rank}. Specifically, a small re-ranking model is deployed on each mobile device to capture the user's real-time interest shifts and re-rank unexposed items accordingly. The small on-device re-ranking model comprises the embedding vectors of candidate items, generated by the cloud-based large ranking model, along with additional network components, such as attention layers and multilayer perceptrons (MLPs). Rather than deploying the full embedding table, which is common in cloud settings but infeasible on resource-constrained mobile devices, a subset of required embeddings is selected as a submodel \cite{proc:mobicom20:submodel,proc:cikm20:edgerec} and dynamically delivered to mobile devices for efficient deployment.

\subsubsection{On-Device Small Unimodal Model and Cloud-Based Large Multimodal Model Collaboration in Livestreaming}

To analyze livestreaming content, such as identifying which item a streamer is promoting, the typical workflow involves each mobile device uploading video frames along with audio or text transcriptions obtained through automatic speech recognition (ASR) to the cloud, where a large multimodal model processes the data. However, the key bottleneck is the heavy workload of cloud-based multimodal analysis, which limits coverage to only the most popular streamers and part of video frames. Taobao Live thus proposed to deploy a small unimodal (i.e., visual-only) model directly on each mobile device \cite{proc:icdm23:livestream}. This on-device unimodal model performs feature extraction and pre-recognition. Only frames that cannot be recognized with high confidences locally will trigger an upload of their unimodal features to the cloud, where a large multimodal model handles recognition without the need for additional feature extraction. To adapt to each streamer's personalized and dynamic content, the unimodal model is incrementally trained on local samples directly on the mobile device. Since these personalized unimodal features are not aligned with the features in other modalities on the cloud, they are transformed into prompts and inserted as input prefixes to the multimodal model, instructing multimodal fusion and analysis in a streamer-specific manner.

\subsubsection{Cloud-Based LLM and On-Device Small LLM Collaboration for Personal Intelligent Assistant} Smartphone manufacturers have been exploring deploying a family of generative models with different parameter sizes and capabilities across mobile devices and cloud servers to power personal intelligent services. A prominent example is Apple Intelligence \cite{arxiv24:apple:intelligence,link:apple:intell}, which integrates on-device small models, cloud-based larger models, and OpenAI's ChatGPT to to deliver efficient and contextually relevant assistance, while upholding the commitment to user privacy. The series of Apple's self-developed models have been fine-tuned for users’ everyday tasks, such as writing and refining text, prioritizing and summarizing notifications, creating playful images for conversations with family and friends, and taking in-APP actions to simplify interactions across APPs. The orchestration layer determines the appropriate model based on task complexity and privacy considerations. It prioritizes on-device small LLM processing, then server-based larger LLM, and resorts to ChatGPT only when necessary and with user approval. In particular, on-device LLM is in the parameter size of roughly 3 billion and embedded directly into iPhone, iPad, and Mac, enabling fast and private handling of simple tasks such as text suggestions and lightweight image processing. The server-based LLM is larger and available with Private Cloud Compute platform and running on Apple silicon servers without retaining or exposing user data, supporting more complex tasks that require greater computational power. In instances where tasks surpass both on-device and server-based LLMs, ChatGPT will be invoked, to handle complex queries, such as intricate questions about photos or documents and creative content generation. Importantly, mobile device users control when ChatGPT is used and will be asked before any information is shared.

\section{Future Directions}\label{sec:future:dir}

Despite advances in both academia and industry, several open problems remain across the system, algorithm, and application layers. We point out some potential directions for future study.

At the system layer, one important direction is {\em to address the tension between universality and specialization}. For example, mainstream deep learning engines were originally designed and optimized for small discriminative models. Supporting Transformer-based large autoregressive models now poses a key challenge regarding whether to adapt existing engines or develop specialized ones. When further factoring in the heterogeneous hardware and software across mobile devices and cloud servers, developing a universal engine that supports both on-device small models and cloud-based large models, or alternatively, designing platform-specific and/or model-specific engines, remains an open challenge. The development and maintenance workload, practical deployment requirement, and model execution performance (e.g., latency, memory usage, power consumption) should be carefully considered. The second important direction is {\em the co-design of system and algorithm for better collaboration between on-device small models and cloud-based large models}. For example, data-based, feature-based, and parameter-based collaboration frameworks, as well as interaction paradigms ranging from single-device-to-cloud to multi-device-to-cloud, impose different requirements on the underlying system. Specifically, compared with parameter-based collaboration framework, data-based collaboration framework tends to impose additional requirements, including support for fine-grained data processing, database indexing and retrieval, sample lifecycle management, and privacy preservation, on mobile devices and cloud servers. In contrast, parameter-based collaboration normally demands efficient support for on-device training. In addition, compared with single-device-to-cloud interaction, the multi-device-to-cloud paradigm presents increased system-level challenges, particularly in managing synchronization or asynchronization among mobile devices that may be intermittently connected and exhibit variations in model execution speed. The third direction is to {\em build large-scale, real-device platforms that advance academic research and streamline task development and testing in industry}. Most existing research relied on server-based simulations to evaluate proposed designs. However, such simulations often fall short in capturing the practical hardware and software constraints of mobile devices when executing small models, as well as the real-world communication dynamics involved in large and small models collaboration between mobile devices and cloud servers. As a result, practical efficiency metrics essential for real-world deployment were often overlooked or insufficiently evaluated. In contrast, industrial workflows typically adopt a single-device-to-cloud interaction paradigm, conduct limited testing on a few real mobile devices, and then deploy at scale through commercial mobile APPs. Enabling large-scale, real-device testing for more complex, multi-device-to-cloud interaction paradigms remains a challenging problem.

At the model and learning algorithm layer, the first fundamental direction is to {\em establish a theoretical analysis framework and provide strict theoretical guarantees for different collaborative learning algorithms and different pairs of on-device small model and cloud-based large model}. The optimization objective of on-device small model is to minimize the loss over the the corresponding user's local data distribution, whereas the optimization objective of cloud-based large model is to minimize the loss over the global data, which are normally a mixture of many mobile devices' local data and may also contain some other features that are not available on mobile devices. In practice, some users may refuse to share local data or part of features with the cloud. In addition, on-device small model and cloud-based large model interact through data, feature, and parameter for joint optimization. How to analyze collaborative learning of on-device small model and cloud-based large model in theory and further guide the model choice and the learning algorithm design is an important problem. The second important direction is to {\em design new collaborative learning frameworks specific to the characteristics of on-device small model and cloud-based large model.} The key advantage of on-device small model is its presence at the application scenario, data source, and users, enabling low response latency, personalized adaptation to users’ unique and dynamic characteristics, enhanced data privacy, and lower operational cost. In contrast, the cloud-based large model excels in its ability to generalize across diverse tasks and domains. In addition, on-device small model and cloud-based large model may differ significantly in their functionalities and architectures (e.g., a small CNN and a LLM). Designing a new collaborative learning framework that seamlessly integrates the strengths of both on-device small model and cloud-based large model while incorporating their specificity to achieve greater effectiveness and efficiency than either model alone remains an important problem for research. One potential way is to model task-level collaborative learning using a multi-agent framework \cite{proc:nips23:CAMEL,proc:acl24:chatdev,proc:iclr24:agentverse}, where the key distinction is that on-device small model and cloud-based large model are not strictly a weak agent and a strong agent or an external tool and an agent; and meanwhile, agents are dynamic and capable of evolving, such as by updating their model parameters. The third direction is to {\em reconsider the design of on-device small model in the context of collaborative learning with cloud-based large model}. Given the presence of a large pretrained model on the cloud, an important question is how to design the architecture of the on-device small model and determine an effective training strategy, so that it satisfies both task requirements on mobile devices and resource constraints. It is necessary to analyze the conditions under which on-device small model should be built from scratch versus derived through compression and distillation from cloud-based large model. Further considering the assistance of the cloud-based large model within the collaborative learning framework, rather than relying solely on the on-device small model as in most existing work, it is important to investigate how to make the on-device small model more efficient and effective. For example, a series of LLMs with varying sizes may need to be rearchitected to better suit the collaborative learning context. The fourth direction is to {\em develop personalized learning algorithms for on-device small model, while preserving generalization capability with the assistance of the cloud-based model.} One major challenge in enhancing on-device small model is the limited availability of abundant, high-quality, labeled data on each individual mobile device. Achieving significant improvements over a unified model remains difficult in the context of on-device model personalization. Furthermore, enhancing on-device model often faces the dilemma of balancing personalization with generalization. Effectively incorporating assistance from a cloud-based large model requires careful design. Another critical challenge lies in the resource constraints of mobile devices. For example, tuning billion-scale LLMs can easily lead to out-of-memory errors. Thus, designing efficient and lightweight on-device enhancement algorithms is essential. The fifth direction is to {\em enhance the cloud-based large model through privacy-preserving use of on-device heterogeneous user data.} In industrial applications, on-device user data are frequently subject to anonymity constraints, with privacy-sensitive fields withheld from upload. These limitations pose a substantial challenge to enabling cloud-based large models to recognize personalized user features and perform effective model updates. Additionally, with data across different mobile devices often following non-iid, a key challenge is how the cloud-based large model can exploit these data to deliver more effective and personalized services, and what role on-device small models play in supporting this objective. The last but not least direction is to {\em extend the collaborative learning of on-device small model and cloud-based large model to more ubiquitous settings.} An interesting topic is to introduce edge servers equipped with medium-sized models, extending the traditional two-layer architecture into a three-layer architecture. This extension could enable more effective and efficient collaborative learning, while also introducing new open and challenging research questions. Looking ahead, a compelling vision is to orchestrate models of varying scales, architectures, and capabilities across vast networks of heterogeneous devices to support collaborative inference and continual learning.

At the application layer, the future direction lies in {\em deploying collaborative learning of on-device small model and cloud-based large model in more industrial scenarios to meet practical requirements}. Potential applications to be put into large-scale production use include, but are not limited to, productivity tools on mobile devices, in-vehicle digital assistants, task planning and execution in robotics through the coordination of large and small ``brains'', automated operations and maintenance in internet of things (IoT), generative search engines, video surveillance, as well as intelligent interaction and rendering in XR, among others. 

\section{Conclusion}

Collaborative learning between on-device small model and cloud-based large model offers a promising new paradigm that balances personalization, efficiency, and privacy. Unlike traditional cloud-centric learning framework that depends heavily on centralized servers for both training and inference, this collaborative framework enables a distributed splitting of learning task: the small model on mobile device handles local, real-time tasks, while the large model on the cloud is responsible for global reasoning and complex computations. This framework not only reduces latency and operational costs but also enables the delivery of user-aware, adaptive intelligence across diverse application domains. The rapid momentum in this area is evident from the growing number of contributions spanning hardware acceleration, system and learning engine optimization, model architecture, algorithmic innovation, and real-world deployment.

Through this survey, we have provided a structured overview of the foundational layers of this emerging paradigm, highlighting key advances and ongoing challenges. We emphasize that the problems discussed herein, though not exhaustive, reflect both the current state and the trajectory of the field as seen through the lens of recent academic and industrial efforts. Notably, many challenges at different layers as listed in Section \ref{sec:key:problem} and Section \ref{sec:future:dir} remain underexplored. We hope this survey will serve as a reference point and a catalyst for future research on the co-evolution of cloud-based intelligence and on-device adaptability.

\begin{acks}
This work was supported in part by National Key R\&D Program of China (No. 2022ZD0119100), China NSF grant No. 62025204, No. 62202296, No. 62272293, No. 62441236, and No. U24A20326, Alibaba Innovation Research (AIR) Program, SJTU-Huawei Research Program, and Tencent WeChat Research Program. The opinions, findings, conclusions, and recommendations expressed in this paper are those of the authors and do not necessarily reflect the views of the funding agencies or the government.
\end{acks}

\bibliographystyle{ACM-Reference-Format}
\bibliography{ref,kd-ref}


\end{document}